\pdfoutput=1

\documentclass[11pt]{article}

\usepackage{emnlp2021}

\usepackage{times}
\usepackage{latexsym}
\usepackage{mathtools}
\usepackage{amssymb}

\usepackage[T1]{fontenc}

\usepackage[utf8]{inputenc}

\usepackage{microtype}

\usepackage{comment}

\title{An Empirical Study on Leveraging Position Embeddings for Target-oriented Opinion Words Extraction}

\author{Samuel Mensah \\
  Computer Science Department \\
  University of Sheffield, UK \\
  \texttt{\small s.mensah@sheffield.ac.uk} \\\And
  Kai Sun \\
  BDBC and SKLSDE \\
  Beihang University, China \\
  \texttt{\small sunkai@buaa.edu.cn} \\\And
  Nikolaos Aletras \\
  Computer Science Department \\
  University of Sheffield, UK \\
  \texttt{\small n.aletras@sheffield.ac.uk} \\}

\begin{document}
\maketitle

\begin{abstract}

Target-oriented opinion words extraction (TOWE) \cite{fan2019target} is a new subtask of target-oriented sentiment analysis that aims to extract opinion words for a given aspect in text. Current state-of-the-art methods leverage position embeddings to capture the relative position of a word to the target. However, the performance of these methods depends on the ability to incorporate this information into word representations. In this paper, we explore a variety of text encoders based on pretrained word embeddings or language models that leverage part-of-speech and position embeddings, aiming to examine the actual contribution of each component in TOWE. We also adapt a graph convolutional network (GCN) to enhance word representations by incorporating syntactic information. Our experimental results demonstrate that BiLSTM-based models can effectively encode position information into word representations while using a GCN only achieves marginal gains. Interestingly, our simple methods outperform several state-of-the-art complex neural structures.

\end{abstract}



\section{Introduction}

Target-oriented opinion words extraction (TOWE) ~\cite{fan2019target} is a fine-grained task of target-oriented sentiment analysis~\cite{liu2012sentiment} aiming to extract opinion words with respect to an opinion target (or aspect) in text. Given the sentence {\it ``The \underline{food} is good but the \underline{service} is extremely slow''}, TOWE attempts to identify the opinion words {\it ``good''} and {\it ``extremely slow''} corresponding respectively to the targets {\it ``food''} and {\it ``service''}.  
TOWE is usually treated as a sequence labeling problem using the BIO tagging scheme~\cite{ramshaw1999text} to distinguish the {\bf B}eginning, {\bf I}nside and {\bf O}utside of a span of opinion words. Table \ref{tab:example-prediction} shows an example of applying the BIO tagging scheme for TOWE.

\begin{table}[!tbp]
    \centering
    \small
    \begin{tabular}{l} \hline
      {\bf Sentence:} \\ \hline \hline
      The \underline{food} is good but the \underline{service} is extremely slow. \\ \hline
      {\bf True Labels for target `food':} \\ \hline \hline
      The/{\bf O} \underline{food}/{\bf O} is/{\bf O} good/{\bf B} but/{\bf O} the/{\bf O} service/{\bf O}\\ is/{\bf O} extremely/{\bf O} slow/{\bf O}.     \\ \hline
      {\bf True Labels for target `service':} \\ \hline \hline
      The/{\bf O} food/{\bf O} is/{\bf O} good/{\bf O} but/{\bf O} the/{\bf O} \underline{service}/{\bf O}\\ is/{\bf O} extremely/{\bf B} slow/{\bf I}.     \\ \hline 
      {\bf TOWE Extraction Results:} \\ \hline \hline
      \{(food, good), (service, extremely slow)\} \\ \hline
    \end{tabular}
    \caption{Identifying target-oriented opinion words in a sentence. Underlined words are opinion targets. Spans tagged {\bf B} and {\bf I} are considered as opinion words.}
    \label{tab:example-prediction}
\end{table}

Learning effective word representations is a critical step towards tackling TOWE. Traditional work \cite{zhuang2006movie,hu2004mining,qiu2011opinion} has used hand-crafted features to represent words which do not often generalize easily. More recent work \cite{DBLP:conf/emnlp/LiuJM15,fan2019target,DBLP:conf/aaai/WuZDHC20,DBLP:conf/emnlp/VeysehNDDN20} has explored neural networks to learn word representations automatically. 

Previous neural-based methods \cite{DBLP:conf/emnlp/LiuJM15,fan2019target} has used word embeddings \cite{collobert2008unified,mikolov2013distributed,pennington2014glove} to represent the input. However, TOWE is a complex task that requires a model to know the relative position of each word to the aspect in text. Words that are relatively closer to the target usually express the sentiment towards that aspect \cite{zhou2020position}. 

%

%




\citet{fan2019target} employ Long Short-Term Memory (LSTM) networks \cite{hochreiter1997long} to encode the target position information in word embeddings. \citet{DBLP:conf/aaai/WuZDHC20} transfer latent opinion knowledge into a Bidirectional LSTM (BiLSTM) network that leverages word and position embeddings \cite{zeng2014relation}. Recently, \citet{DBLP:conf/emnlp/VeysehNDDN20} have proposed ONG, a method that combines BERT (Bidirectional Encoder Representations from Transformers) \cite{devlin2018bert}, position embeddings, Ordered Neurons LSTM (ON-LSTM) \cite{shen2018ordered},\footnote{An LSTM variant with an inductive bias toward learning latent tree structures in sequences.} and a graph convolutional network (GCN) \cite{kipf2016semi} to introduce syntactic information into word representations. While this model achieves state-of-the-art results, previous studies have shown that the ON-LSTM does not actually perform much better than LSTMs in recovering latent tree structures \cite{dyer2019critical}. Besides, ON-LSTMs perform worse than LSTMs in capturing short-term dependencies \cite{shen2018ordered}. Since opinion words are usually close to targets in text, ON-LSTM risks missing the relationship between the aspect and any information (e.g. position) relating to the opinion words.

In this paper, we empirically evaluate a battery of popular text encoders which apart from words, take positional and part-of-speech information into account. Surprisingly, we show that methods based on BiLSTMs can effectively leverage position embeddings to achieve competitive if not better results than more complex methods such as ONG on standard TOWE datasets. 
Interestingly, combining a BiLSTM encoder with a GCN to explicitly capture syntactic information achieves only minor gains. This empirically highlights that BiLSTM-based methods have an inductive bias appropriate for the TOWE task, making a GCN less important.

\section{Methodology}
Given sentence $s=\{w_1,\ldots,w_n\}$ with aspect $w_t\in s$, our approach consists of a text encoder that takes as input a combination of words, part-of-speech and position information for TOWE. We further explore enhancing text encoding by incorporating information from a syntactic parse of the sentence through a GCN encoder.




\subsection{Input Representation}

\paragraph{Word Embeddings:}
We experiment with Glove word vectors  \cite{pennington2014glove} as well as BERT-based representations, extracted from the last layer of a BERT base model \cite{devlin2018bert} fine-tuned on TOWE.

\paragraph{Position Embeddings (POSN):} We compute the relative distance $d_i$ from $w_i$ to $w_t$ (i.e., $d_i=i-t$), and lookup their embedding in a randomly initialized position embedding table. 

\paragraph{Par-of-Speech Tag Embeddings (POST):} We assign part-of-speech tags to each word token using the Stanford parser,\footnote{https://stanfordnlp.github.io/CoreNLP/} and lookup their embedding in a randomly initialized POST embedding table.

\paragraph{Combined Input:} We consider two types of input representations: 

\begin{enumerate}
    \item {\bf Glove Input (G)}: Constructed from  concatenating Glove word embeddings, POST and POSN embeddings for each token.
    
    \item {\bf BERT Input (B)}: Constructed from concatenating BERT vectors with POSN embeddings for each word token following a similar approach as \cite{DBLP:conf/emnlp/VeysehNDDN20}.\footnote{Early experimentation with RoBERTa \cite{liu2019roberta} yielded lower performance.} We ignore POST embeddings since BERT is efficient in modeling such information \cite{tenney2019you}. 

\end{enumerate}





\subsection{Text Encoders}
We experiment with the following neural encoders that take word vector representations as input:


\paragraph{CNN:} A single layer convolutional neural network \cite{lecun1990handwritten}. Given a word $w_i\in s$, the CNN takes a fixed window of words around it and applies a filter on their representation to extract a feature vector for $w_i$. We concatenate the feature vectors corresponding to different filters for $w_i$ to compute word representations.

\paragraph{Transformer:} A Transformer encoder \cite{vaswani2017attention} that takes a linear transformation of the input words to learn contextualized representations.

\paragraph{BiLSTM:} A bi-directional LSTM that takes the input representation and models the context in a forward and backward direction.

\paragraph{ON-LSTM:} A variant of the LSTM neural network proposed by \cite{shen2018ordered} which has an inductive bias toward learning latent tree structures.

\subsection{GCN Encoder}

First, we interpret the syntactic parse tree as an adjacency binary matrix $A^{n\times n}$ ($n$ is the sentence length) with entries $A_{ij}=1$ if there is a connection between nodes $i$ and $j$, and $A_{ij}=0$ otherwise. To apply a GCN on $A$, we consider the tree with self-loops at each node (i.e., $A_{ii}=1$), ensuring nodes are informed by their corresponding representations at previous layers. Formally, let $H^{(k)}$ be the output at the $k$-th GCN layer, $H^{(k)}$ is given by:
\begin{equation}
H^{(k)} = {\rm ReLU}(AH^{(k-1)}W^{(k)}  ) + H^{(k-1)}
\label{eqn:gcn}
\end{equation}
\noindent where $k=1,\ldots,K$, $W^{(k)}$ is a parameter matrix at layer $k$. $RELU$ is used as the activation function.  $H^{(0)}$ corresponds to the set of word representations extracted by the text encoder. The second term in \eqref{eqn:gcn} induces a residual connection that retains the contextual information of $H^{(0)}$ during the propagation process \cite{sun2020relation}. 


\subsection{Classification and Optimization}


Our model uses the representation $H^{(l)}$ (where $l\geq 0$), applies a linear layer and then normalize it with a softmax function to output a probability distribution over the set \{B,I,O\} for each word in the input. During training, we minimize the cross-entropy function for each word in text for the entire training set.





\section{Experiments and Results}

\subsection{Baselines}
We compare our methods with Distance-rule \cite{DBLP:conf/kdd/HuL04}; Dependency-rule \cite{DBLP:conf/cikm/ZhuangJZ06}; LSTM$_{\rm word}$ and BiLSTM$_{\rm word}$ \cite{DBLP:conf/emnlp/LiuJM15}; Pipeline \cite{fan2019target}; TC-BiLSTM \cite{fan2019target}; IOG \cite{fan2019target}; LOTN \cite{DBLP:conf/aaai/WuZDHC20}; and ONG \cite{DBLP:conf/emnlp/VeysehNDDN20}.\footnote{Note that LSTM$_{\rm word}$/BiLSTM$_{\rm word}$ only use word embeddings as input.} 

\subsection{Data}
Following~\cite{wu2020latent}, we use four benchmark datasets including restaurant (Res14, Res15, Res16) and laptop (Lap14) reviews from Semeval~\cite{pontiki2014semeval,pontiki2015semeval,pontiki2016semeval}.  We use the preprocessed data provided by~\citet{DBLP:conf/naacl/FanWDHC19}. Table~\ref{tab:data_stat_sentiment} shows the dataset statistics.


\begin{table}[!htbp]
 \tiny
    \centering
\begin{tabular}{l|c|c|c|c|c|c}\hline
Dataset& \#Sent. & \#ASL  &\#AT  & \#OT & \#D.Dist. &\#S.Dist.\\   \hline
Lap14 (Train)   &1151 &20.78 &1632 &1877 & 2.40 &4.25\\
Lap14 (Test)    &343  &17.33 &482 &567 & 2.03 &4.00\\
\hline
Res14 (Train)  &1625 &19.11 &2636 &3057 &2.11 &3.68\\ 
Res14 (Test)  &500 &19.22 &862 &1028 &2.01 &3.97\\
\hline 
Rest15 (Train) &754 &16.50 &1076 &1277 &1.97 &3.62\\
Rest15 (Test) &325 &17.47 &436 &493 &2.13 &3.53\\
 \hline
Rest16 (Train) &1079 &16.78 &1512 &1770 &2.01 &3.59\\
Rest16 (Test) &328 &16.54 &456 &524 &1.93 &3.43\\
 \hline
\end{tabular}
    \caption{Dataset Statistics. No. of sentences (\#Sent), Avg. sentence length (\#ASL), No. of aspect terms (\#AT), No. of opinion words (\#OT), Avg. dependency distance (\#D.Dist) and Avg. sequential distance (\#S.Dist) between aspect and opinion.}
    \label{tab:data_stat_sentiment}
\end{table}



\subsection{Implementation Details}

Hyper-parameters are tuned on 20$\%$ of samples randomly selected from the train set since there is no development set.\footnote{We use 300-dim Glove word vectors \cite{pennington2014glove} and apply a dropout of 0.8. Dimensions of part-of-speech and position embeddings are set to 30, but dimensions of position embeddings for pretrained models are set to 100. The CNN uses three filters with sizes 3, 4 and 5 and has a hidden dimension of 300. All other models have a hidden dimension of 200. The number of GCN layers is set over $K\in\{1,\ldots,5\}$. Experiments are performed on NVIDIA Tesla V100.} We use the Adam optimizer to train all models. Models that use Glove word vectors are optimized with learning rate $1e^{-3}$ and trained for 100 epochs with batch size 16. Models that use BERT hidden vectors are optimized with learning rate $1e^{-5}$ and trained with batch size 6. Our source code is publicly available.\footnote{\url{https://github.com/samensah/Encoders_TOWE_EMNLP2021}} 




\begin{table*}[h!]
\centering
\scriptsize
\begin{tabular}{l|ccc|ccc|ccc|ccc|c}\hline
&\multicolumn{3}{c|}{Lap14} &
\multicolumn{3}{c|}{Res14} &
\multicolumn{3}{c|}{Res15} &
\multicolumn{3}{c}{Res16}  
\\   \cline{2-14}
{Model} & Prec & Rec & F1 & Prec & Rec & F1 & Prec & Rec & F1 & Prec & Rec & F1 & Avg.F1 \\ \hline %
 Distance-rule &50.13 &33.86 &40.42 &58.39 &43.59 &49.92 &54.12 &39.96 &45.97 &61.90 &44.57 &51.83 & 47.04 \\ %
Dependency-rule &45.09 &31.57 &37.14 &64.57 &52.72 &58.04 &65.49 &48.88 &55.98 &76.03 &56.19 &64.62 & 53.95 \\ %
LSTM$_{\rm word}$ &55.71 &57.53 &56.52 &52.64 &65.47 &58.34 &57.27 &60.69 &58.93 &62.46 &68.72 &65.33 & 59.78 \\ %
BiLSTM$_{\rm word}$ &64.52 &61.45 &62.71 &58.34 &61.73 &59.95 &60.46 &63.65 &62.00 &68.68 &70.51 &69.57 & 63.56 \\ %
Pipeline &72.58 &56.97 &63.83 &77.72 &62.33 &69.18 &74.75 &60.65 &66.97 &81.46 &67.81 &74.01 & 68.50  \\ %
TC-BiLSTM &62.45 &60.14 &61.21 &67.65 &67.67 &67.61 &66.06 &60.16 &62.94 &73.46 &72.88 &73.10 & 66.22  \\ %
IOG & 73.24 & 69.63 &71.35 &82.85 &77.38 &80.02 &76.06 &70.71 &73.25 &82.25 &78.51 &81.69 & 76.58 \\ %
LOTN &  77.08 &67.62 &72.02 &84.00 &80.52 &82.21 &76.61 &70.29 &73.29 &86.57 &80.89 &83.62 & 77.79 \\ %
ONG & 73.87 &  77.78 & 75.77  &83.23 &81.46 &82.33 &76.63 & {  81.14} &78.81 &87.72 &  84.38  &86.01 & 80.73  \\ %


 \hline
 \multicolumn{13}{l}{\bf Glove Input}\\

 \hline

Transformer(G) & 68.33 &61.91 &64.91 &71.77 &70.29 &70.98 &78.90 &59.07 &67.41 &83.59 &70.57 &76.49 &69.94 \\ 
CNN(G) & 64.81 &73.83 &69.00 &75.86 &78.83 &77.29 &68.21 &73.87 &70.91 &76.93 &84.77 &80.64 &74.46 \\ 
ON-LSTM(G)  &69.27 &69.70 &69.47 &83.01 &76.98 &79.87 &76.19 &74.24 &75.20 &84.17 &82.90 &83.52 &77.02\\
BiLSTM(G)  &76.49 &70.94 &73.59 &86.22 &83.44 &84.80 &81.49 &77.93 &79.66 &88.96 &84.05 &  87.36 &81.35 \\ \hline 
Transformer+GCN(G)  &66.32 &70.83 &68.45 &82.98 &75.14 &78.82 &76.80 &69.45 &72.88 &84.71 &79.92 &82.25 &75.60 \\ 

CNN+GCN(G)  &66.88 &74.88 &70.65 &82.45 &80.12 &81.24 &75.32 &73.75 &74.51 &82.17 &84.89 &83.48 &77.47\\ 


ON-LSTM+GCN(G) &71.63 &74.04 &72.75 &87.06 &80.97 &83.90 &80.18 &77.53 &78.83 &89.89 &83.97 &86.82 &80.58 \\
BiLSTM+GCN(G)  & 76.49 &74.46 & 75.46 & 87.60 &83.66 & 85.57 &82.32 &78.82  & 80.52 & 91.63 &85.65 &  {\bf 88.52} & 82.52 \\ 

 \hline
 \multicolumn{13}{l}{\bf BERT Input}\\

 \hline

 {Transformer(B)} &78.88 &78.03 &78.13 &83.97 &84.40 &84.18 &82.37 &78.21 &80.22 &88.22 &84.05 &86.06 &82.14 \\
 {CNN(B)}  &77.94 &75.91 &76.87 &86.35 &82.16 &84.20 &80.01 &78.62 &79.30 &88.50 &82.41 &85.33 &81.43 \\
 {ON-LSTM(B)} &77.96 &77.53 &77.71 &85.58 &83.25 &84.39 &82.57 &78.34 &80.38 &87.76 &83.55 &86.54 &82.26\\
BiLSTM(B) & 78.38 &  78.27 & {78.25} & 86.38 & {84.82} &  85.60 &  82.17 & 78.78 &  80.41 &  89.94 & 84.16 &86.94 &  82.80 \\ \hline 

 {Transformer+GCN(B)} &79.38 &77.04 &78.19 &85.43 &84.18 &84.79 &82.21 &79.55 &{\bf 80.84} &89.34 &84.16 &86.66 &82.62 \\
 {CNN+GCN(B)} &79.19 &76.19 &77.62 &84.96 &84.08 &84.50 &82.39 &77.36 &79.77 &88.16 &84.09 &86.06 &81.98 \\
 {ON-LSTM+GCN(B)} &80.33 &76.01 &77.96 &85.68 &84.03 &84.83 &82.14 &78.18 &80.07 &89.35 &83.93 &86.54 &82.35 \\
BiLSTM+GCN(B) &  79.72 & {78.06} &  {\bf 78.82} & 86.45 &  85.06 &  {\bf 85.74} &  83.37 & 77.93 &80.54 & 88.98 &  {85.80} & {87.35} & {\bf 83.11} \\ 



\end{tabular}
\caption{Results of experiments across baseline methods (across 5 runs). Results of compared models are retrieved from~\cite{DBLP:conf/emnlp/VeysehNDDN20}. The best F1 performance is bold-typed.}
\label{table:aspect-oriented}
\end{table*}

\begin{table*}[h!]
\centering
\scriptsize
\begin{tabular}{l|ccc|ccc|ccc|ccc}\hline
&\multicolumn{3}{c|}{Lap14} &
\multicolumn{3}{c|}{Res14} &
\multicolumn{3}{c|}{Res15} &
\multicolumn{3}{c}{Res16}
\\   \cline{2-13}
{Model} & Prec & Rec & F1 & Prec & Rec & F1 & Prec & Rec & F1 & Prec & Rec & F1\\ \hline
{\bf BiLSTM+GCN(G)} & 76.49 &74.46 &75.46 & 87.60 & 83.66 & 85.57 & 82.32 &78.82 & 80.52 & 91.63 & 85.65 & 88.52 \\ \hline
\quad --- GCN &76.49 &70.94 &73.59 &86.22 &83.44 &84.80 &81.49 &77.93 &79.66 &88.96 &84.05 &87.36\\
\quad --- GCN, POST &75.38 &70.12 &72.63 &86.83 &82.94 &84.83 &82.45 &75.58 &78.85 &88.71 &84.01 &86.29 \\
\quad ---  GCN, POST, POSN &61.65 &62.08 &61.80 &63.17 &56.63 &59.66 &62.16 &61.54 &61.76 &70.11 &70.23 &70.08 \\
\hline
{\bf BiLSTM+GCN(B)} &  79.72 & {78.06} &  { 78.82} & 86.45 &  85.06 &  { 85.74} &  83.37 & 77.93 &    80.54 & 88.98 &  {85.80} & { 87.35} \\ \hline 
\quad --- GCN  & 78.38 &  78.27 & {78.25} & 86.38 & {84.82} &  85.60 &  82.17 & 78.78 &  80.41 &  89.94 & 84.16 &86.94  \\ 
\quad --- GCN, POSN &62.92 &72.17 &67.21 &60.84 &64.42 &62.54 &63.88 &64.42 &63.97 &69.59 &71.45 &70.39 \\
\end{tabular}
\caption{Precision, Recall and F1 scores of ablated models on the benchmark datasets (across $5$ runs).}
\label{table:ablation_study}
\end{table*}


\subsection{Performance Comparison}

Table~\ref{table:aspect-oriented} presents the results of all methods. Our models that use Glove Input (or BERT Input) are appended with ``G''(or ``B'') to distinguish them. We report precision (Prec), recall (Rec), F1 score and average F1 score (Avg.F1) across all datasets. 

\paragraph{Comparison of Text Encoders:} 
We first observe that CNN(G) is adept at exploiting the information from simpler word representations (Glove), outperforming the Transformer(G) by +4.52 Avg.F1. We believe that this behavior is due to the fact that TOWE is a short-sequence task (see \#ASL in Table \ref{tab:data_stat_sentiment}). This assumption lies well with previous observations by \cite{yin2021neural}, which found that CNNs often perform better than Transformers at short-sequence tasks. However, the Transformer(B) is able to improve performance and even outperform CNN(B) by +0.71 Avg.F1 by using BERT.

In addition, we find that ON-LSTM(G) and ON-LSTM(B) lag behind BiLSTM(G) and BiLSTM(B) by 4.33 and 0.54 Avg.F1 respectively. ON-LSTM performs worse than LSTMs on tasks that require tracking short-term dependencies \cite{shen2018ordered}. Since opinion words are usually close to the target in the sequence (see \#ASL vrs. \#S.Dist. in Table~\ref{tab:data_stat_sentiment}), tracking short-term dependency information is important in TOWE. This explains why BiLSTM(G)(or BiLSTM(B)) achieves a better performance over ON-LSTM(G)(or ON-LSTM(B)). 

The performance of BiLSTM(G) over BiLSTM$_{\rm word}$ suggests that the substantial boost in performance comes from either part-of-speech or position embeddings. We later perform an ablation experiment to examine which information is more useful. Interestingly, BiLSTM(G) outperforms the current state-of-the-art ONG by +0.62 Avg.F1 despite its simple architecture, demonstrating the importance to first experiment with simpler methods before designing more complex structures. 

\paragraph{Comparison of Text+GCN Encoders:} Adding a GCN over any text encoder generally improves performance. This happens because the GCN provides additional syntactic information that is helpful for representation learning. We find that BiLSTM+GCN(G) achieves few gains over BiLSTM(G) while other text encoders including Transformer+GCN(G) and CNN+GCN(G) achieve relatively higher gains than their counterparts. This suggest that BiLSTM(G) has an inductive bias appropriate for the TOWE task and the performance mostly depends on the quality of the input representation. We observe that when using BERT embeddings, there is a minimal performance difference between using GCNs or not. We attribute this to the expressiveness of BERT embeddings and its ability to capture syntactic dependencies \cite{jawahar2019does}. 
Overall results suggest that our proposed method outperforms SOTA consistently across datasets.

\subsection{Ablation Study}
We perform ablation experiments on the two best performing models, BiLSTM+GCN(G) and BiLSTM+GCN(B), to study the contribution of their different components. The results are shown in Table~\ref{table:ablation_study}. On BiLSTM+GCN(G), as we consecutively remove the GCN and POST embeddings from the input representation, we observe a slight drop in performance. The results indicate that POST embeddings as well as the GCN are not critical components for BiLSTM+GCN(G). Therefore, they can be ignored to reduce model complexity. However, we observe a substantial drop in performance by removing the position embedding from the input representation, obtaining an F1 score equivalent to BiLSTM$_{\rm word}$ across datasets. Similarly, removing the position embeddings in BiLSTM+GCN(B) causes a substantial drop in performance. The results suggest that leveraging position embeddings is crucial for TOWE performance.

\section{Conclusion}


We presented through extensive experiments that by employing a simple BiLSTM architecture that uses input representations from pre-trained word embeddings or language models, POST embeddings and position embeddings, we can obtain competitive, if not better results than the more complex current state-of-the-art methods \citet{DBLP:conf/emnlp/VeysehNDDN20}. The BiLSTM succeeds in exploiting position embeddings to improve  performance. By adapting a GCN to incorporate syntactic information from the sentence we achieve further gains. In future work, we will explore how to improve existing TOWE models by effectively leveraging position embeddings.

\section*{Acknowledgements}
Samuel Mensah and Nikolaos Aletras are supported by a Leverhulme Trust Research Project Grant.

\bibliography{anthology,custom}
\bibliographystyle{acl_natbib}

\end{document}